\documentclass[journal,final,a4paper,twocolumn]{IEEEtran}

\usepackage{scalerel}
\usepackage{tikz}
\usetikzlibrary{svg.path}

\definecolor{orcidlogocol}{HTML}{A6CE39}
\tikzset{
  orcidlogo/.pic={
    \fill[orcidlogocol] svg{M256,128c0,70.7-57.3,128-128,128C57.3,256,0,198.7,0,128C0,57.3,57.3,0,128,0C198.7,0,256,57.3,256,128z};
    \fill[white] svg{M86.3,186.2H70.9V79.1h15.4v48.4V186.2z}
                 svg{M108.9,79.1h41.6c39.6,0,57,28.3,57,53.6c0,27.5-21.5,53.6-56.8,53.6h-41.8V79.1z M124.3,172.4h24.5c34.9,0,42.9-26.5,42.9-39.7c0-21.5-13.7-39.7-43.7-39.7h-23.7V172.4z}
                 svg{M88.7,56.8c0,5.5-4.5,10.1-10.1,10.1c-5.6,0-10.1-4.6-10.1-10.1c0-5.6,4.5-10.1,10.1-10.1C84.2,46.7,88.7,51.3,88.7,56.8z};
  }
}
\newcommand{\codevar}[1]{\texttt{#1}}

\newcommand\orcidicon[1]{\href{https://orcid.org/#1}{\mbox{\scalerel*{
\begin{tikzpicture}[yscale=-1,transform shape]
\pic{orcidlogo};
\end{tikzpicture}
}{|}}}}

\usepackage{times}
\usepackage{epsfig}
\usepackage{graphicx}
\usepackage{amsmath}
\usepackage{amssymb}
\usepackage{subfigure}
\usepackage{array}
\usepackage{url}
\usepackage{amsfonts}
\usepackage{latexsym}
\usepackage{xspace}
\usepackage{bm}
\usepackage{color}
\usepackage{colortbl}
\usepackage{dsfont}
\usepackage{graphics}
\usepackage{multirow}
\usepackage{adjustbox}
\usepackage{tabularx}
\usepackage{ragged2e}
\usepackage{rotating}
\usepackage{tabu}
\usepackage{booktabs}
\usepackage{epstopdf}
\usepackage{multirow}
\usepackage{multicol}
\usepackage{dcolumn}
\usepackage{siunitx}
\usepackage{mathtools}
\usepackage{float}
\usepackage{xcolor,cite,etoolbox}
\usepackage[utf8x]{inputenc}
\usepackage{hyperref}
\hypersetup{hidelinks}
\usepackage{cite}
\usepackage{blindtext}
\usepackage{nicefrac}
\usepackage{float}
\usepackage[colorinlistoftodos]{todonotes}
\usepackage{xcolor}

\usepackage{pifont} 

\newcommand{\best}[1]{\textbf{#1}}
\newcommand{\second}[1]{\underline{#1}}

\begin{document}

\title{HyBiomass: Global Hyperspectral Imagery Benchmark Dataset for Evaluating Geospatial Foundation Models in Forest Aboveground Biomass Estimation}

\author{\IEEEauthorblockN{
Aaron Banze \orcidicon{0009-0007-0783-2968},
Timothée Stassin \orcidicon{0000-0002-1175-4057},
Nassim Ait Ali Braham \orcidicon{0009-0001-3346-3373}, \\
Rıdvan Salih Kuzu \orcidicon{0000-0002-1816-181X}, 
Simon Besnard \orcidicon{0000-0002-1137-103X}, 
and
Michael Schmitt \orcidicon{0000-0002-0575-2362},~\IEEEmembership{Senior Member,~IEEE}}
\thanks{This work has been submitted to the IEEE for possible publication. Copyright may be transferred without notice, after which this version may no longer be accessible. This work is part of the HYPER-AMPLIFAI project funded by the Helmholtz Association of German Research Centres (HGF), contract number ZT-I-PF-4-056.
This work was also supported by the European Space Agency (ESA) as part of the FAST-EO project, under Contract No. 4000143501/23/I-DT. (Corresponding author: \textit{Aaron Banze}.)

A. Banze, N. Ait Ali Braham and R. S. Kuzu are with the Remote Sensing Technology Institute, German Aerospace Center (DLR), 82234 Wessling, Germany (e-mail: \{aaron.banze, nassim.aitalibraham, ridvan.kuzu\}@dlr.de).

T. Stassin and S. Besnard are with the Global Land Monitoring Group, Helmholtz German Research Centre for Geosciences (GFZ), 14473  Potsdam, Germany (e-mail: \{stassin, besnard\}@gfz-potsdam.de).

A. Banze and M. Schmitt are with the Department of Aerospace Engineering, University of the Bundeswehr Munich, 85579 Neubiberg, Germany (e-mail: michael.schmitt@unibw.de).

N. Ait Ali Braham is with the Chair of Data Science in Earth Observation (SiPEO), Technical University of Munich (TUM), 80333 Munich, Germany.

\vspace{-0.34cm}
} 
}

\maketitle

\begin{abstract}
Comprehensive evaluation of geospatial foundation models (Geo-FMs) requires benchmarking across diverse tasks, sensors, and geographic regions. However, most existing benchmark datasets are limited to segmentation or classification tasks, and focus on specific geographic areas.
To address this gap, we introduce a globally distributed dataset for forest aboveground biomass (AGB) estimation, a pixel-wise regression task. This benchmark dataset combines co-located hyperspectral imagery (HSI) from the Environmental Mapping and Analysis Program (EnMAP) satellite and predictions of AGB density estimates derived from the Global Ecosystem Dynamics Investigation lidars, covering seven continental regions.
Our experimental results on this dataset demonstrate that the evaluated Geo-FMs can match or, in some cases, surpass the performance of a baseline U-Net, especially when fine-tuning the encoder.
We also find that the performance difference between the U-Net and Geo-FMs depends on the dataset size for each region and highlight the importance of the token patch size in the Vision Transformer backbone for accurate predictions in pixel-wise regression tasks.
By releasing this globally distributed hyperspectral benchmark dataset, we aim to facilitate the development and evaluation of Geo-FMs for HSI applications. Leveraging this dataset additionally enables research into geographic bias and generalization capacity of \mbox{Geo-FMs}. The dataset and source code will be made publicly available.
\end{abstract}

\begin{IEEEkeywords}
Hyperspectral Imagery, Aboveground Biomass, Geospatial Foundation Models, EnMAP, GEDI, Remote Sensing.
\end{IEEEkeywords}

\vspace{-0.4cm}
\section{Introduction}
\IEEEPARstart{F}{orests} play a vital role in the global carbon cycle, making accurate estimation of related parameters such as aboveground biomass (AGB) of major interest. Optical remote sensing, particularly hyperspectral imagery (HSI), provides detailed spectral information that enables the detection of subtle variations in forest biochemical properties, such as chlorophyll, nitrogen, water content, and lignin. These properties are important indicators of vegetation condition and can serve as proxies for aboveground biomass estimation.
In recent years, the availability of HSI has increased drastically with the launch of satellites such as the Environmental Mapping and Analysis Program (EnMAP) satellite in 2022 \cite{enmap}. This has enabled the development of geospatial foundation models (Geo-FMs) that are pre-trained on HSI and are adapted to the unique requirements of high spectral dimensionality. Initially, Geo-FMs like SatMAE \cite{satmae} were limited to RGB imagery and multispectral imagery (MSI). More recently, several approaches have been developed to address the higher spectral resolution of HSI. SpectralEarth pre-trained models rely on a spectral adapter using 1D convolutions and pooling to reduce the number of channels before the patch embedding layer to be compatible with classical vision backbones \cite{spectralearth}. The Dynamic One-For-All (DOFA) model implements a dynamic hypernetwork to adjust its weights to different wavelengths of any given sensor \cite{dofa}. Lastly, Panopticon utilizes cross-attention to flexibly process varying numbers of spectral bands \cite{panopticon}.

These models have been pre-trained using different datasets. Specifically, the SpectralEarth models were exclusively pre-trained on EnMAP HSI from the SpectralEarth dataset \cite{spectralearth}. Panopticon also utilized the SpectralEarth dataset serving as a basis for HSI training, while relying on additional datasets for other sensor modalities. In contrast, DOFA was pre-trained on the HySpecNet-11k dataset \cite{hyspecnet} as a source for HSI, which is substantially smaller, consisting of 11,483 patches extracted from 250 EnMAP tiles, compared to the 538,974 patches in the SpectralEarth dataset.

Evaluating the performance of Geo-FMs requires comprehensive benchmarking on a diverse range of tasks and sensors, encompassing all regions globally. However, existing benchmark datasets are predominantly limited to segmentation or classification tasks for specific geographic areas \cite{schmitt}. This letter introduces a globally distributed benchmark dataset for the pixel-wise regression task of forest AGB estimation. It consists of co-located HSI from the EnMAP satellite and predictions of aboveground biomass density (AGBD) as reference labels derived from the Global Ecosystem Dynamics Investigation (GEDI) lidar mission. We present the results of benchmarking the aforementioned Geo-FMs adapted to HSI and a baseline U-Net model, which has been shown to achieve great results for forest AGB estimation \cite{unet_pascarella}, using this dataset across seven continental world regions. The regional classification adopted here is consistent with the scheme provided by the GEDI dataset, dividing the earth into seven continental regions \cite{gedi_dubayah}.

\section{Forest Aboveground Biomass \\ Benchmark Dataset}
\begin{figure}[!t]
\centering
\includegraphics[width=\linewidth]{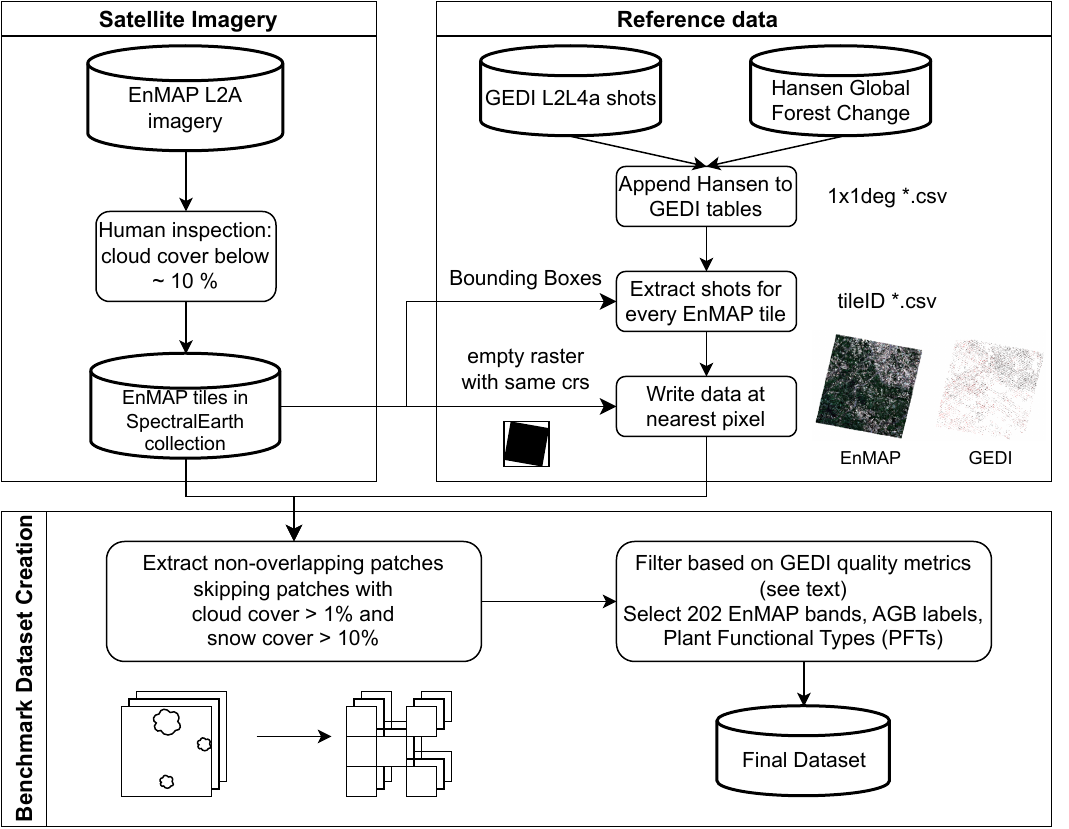}
\vspace{-0.6cm}
\caption{Overview of the data and workflow for creating the benchmark dataset}
\label{fig:data_workflow}
\end{figure}

An overview of the data sources and processing steps used to create the benchmark dataset is provided in \autoref{fig:data_workflow}. In the following subsections we first introduce the EnMAP satellite and outline the criteria we applied to select the tiles included in this dataset. Next, we introduce the GEDI mission specifications and methodology employed to derive the AGBD estimates, which we use as labels. Finally, we describe the procedure for creating the co-located dataset, which consists of spatially non-overlapping patches.

\subsection{EnMAP Hyperspectral Spaceborne Imagery}
The Environmental Mapping and Analysis Program {(EnMAP)} satellite, launched in April 2022, is equipped with a prism-based dual-spectrometer instrument covering the electromagnetic spectrum from visible (420~nm) to shortwave infrared (2450~nm) with 224 spectral bands.
It has a 30~m ground sampling distance, and its scenes cover a 30~km swath width with a length of up to 1000~km per orbit and 5000~km per day.
The satellite is expected to collect data for a period of five years.
In contrast to other missions such as Sentinel-2, EnMAP is a sampling missions, meaning that scenes are acquired according to the user's request or background scientific mission \cite{enmap}.

EnMAP tiles included in this dataset are sourced from Braham et al. \cite{spectralearth}, originally collected for the SpectralEarth dataset. Thereby, tiles from the EnMAP archive from April 2022 to April 2024 were selected based on human visual inspection, keeping the cloud coverage below 10~\%. We removed 22 spectral bands affected by water vapor absorption, resulting in low radiance values, reducing the total number to 202 spectral bands. The same bands have been excluded for the pretraining of the Geo-FMs evaluated in this letter.

\subsection{GEDI-derived AGBD estimates and quality filtering}
The Global Ecosystem Dynamics Investigation (GEDI) mission is designed to provide high-resolution measurements of forest vertical structure using a geodetic-class waveform lidar instrument mounted on the International Space Station (ISS). GEDI employs three lasers to generate eight parallel tracks, each producing footprints of approximately 25 m in diameter, spaced ~60 m apart along-track. Since its launch in 2018,  GEDI has been collecting data across the globe within a latitudinal range of 51.6$^\circ$N to 51.6$^\circ$S. The returned lidar waveforms are used to derive surface elevation, canopy height, and relative height energy metrics, which inform key forest structural attributes. AGBD, expressed in $Mg \times ha^{-1}$, is estimated at the footprint level using parametric models trained on field-based biomass observations. These models are stratified by plant functional type (PFT) and continental region to account for biome-specific structure–biomass relationships \cite{gedi_dubayah}.

The footprint-level GEDI data included in this dataset are sourced from the spatially indexed 1$\times$1 degree tables (EPSG 4326 geographic coordinates) compiled by Burns et al. \cite{gedi_burns}. These data tables were generated by (i) downloading all GEDI L2A, L2B and L4A orbit granule files from April 17 2019 to March 16 2023, (ii) matching those files based on their unique file names, (iii) extracting metrics associated to the default ground-finding algorithm ("\codevar{a0}"), (iv) filtering for highest quality GEDI vegetation shots based on ground elevation and vegetation structure metrics, (v) storing in 1$\times$1 degree tables based on the original shot coordinates.

We applied the L4A high-quality flag developed by Burns et al. \cite{gedi_burns} to filter for shots suitable for AGBD estimates. This involves setting beam sensitivity thresholds depending on the region and PFT, as well as defining a valid range of detected ground elevation among other criteria. For a complete overview of the quality filtering protocol, we refer to the original publication \cite{gedi_burns}.
We further removed GEDI shots with AGBD estimates outside the range [0,~500], as suggested by Carreiras et al. \cite{carreiras}.
Additionally, only GEDI shots captured by power beams were retained, while those originating from coverage beams were excluded, following the recommendation by Lahssini et al. \cite{lahssini}.
To filter for shots where the PFT corresponds to a forest class, we keep GEDI shots with PFT values in range [1,~4], following the 
MODIS MCD12Q1 V006 product Type 5 classification scheme.
We filter the GEDI shots based on PFT, as it is utilized in the preceding step, to select the model for AGBD prediction corresponding to each GEDI shot. However, due to the coarse 500~m spatial resolution of the MODIS product, we additionally filtered based on the Hansen Global Forest Change v1.11 dataset \cite{hansen}, which has a 30~m spatial resolution. Specifically, we utilized the \codevar{landsat\_treecover} variable, which is included in the GEDI data and defines the canopy closure of vegetation taller than 5~m in height in the year 2010, expressed as a percentage per grid cell. Only GEDI shots with \codevar{landsat\_treecover $\geq$~10~\%} were retained.
Finally, we excluded GEDI shots that coincided with gross forest cover loss events occurring between 2010 and 2023 based on the \codevar{lossyear} variable from the Hansen Global Forest Change dataset.

\subsection{Co-locating EnMAP and GEDI data}
The EnMAP and GEDI data were spatially co-located by (i) extracting all GEDI shots from the 1$\times$1 degree tables that intersect the footprint for all EnMAP tiles in our collection and (ii) appending the tabular data of each GEDI shot as additional bands in the raster data at the nearest pixel value based on the geographic coordinates, with no-data values assigned to pixels lacking GEDI data.

The final dataset was created by extracting non-overlapping patches of size 128$\times$128~px from the EnMAP- and GEDI-tiles.
The patchification process utilized local R-Trees to check for overlap with other patches in the same UTM zone, to avoid inaccuracies and distortions that would result from reprojecting to a shared coordinate system. To account for tiles that fall within two UTM zones, a global R-Tree was employed allowing for a maximum overlap area of 10~\% between patches to account for inaccuracies.
During patchification, the tiles were processed in ascending temporal order of their EnMAP acquisition date to minimize the average temporal deviation from the GEDI shots to the EnMAP tile. Patches with cloud cover exceeding 1~\% and snow cover above 10~\% based on the EnMAP quality masks were not included in the final dataset. These threshold values were determined through visual inspection of excluded patches for different threshold values. Considering inherent inaccuracies of the EnMAP quality masks, we found that these thresholds optimize the trade-off between avoiding the exclusion of false positives and ensuring predominantly cloud- and snow-free patches.
Furthermore, patches with less than 1~\% of pixels containing GEDI shots were also omitted.

\begin{figure}[!t]
\centering
\includegraphics[width=\linewidth]{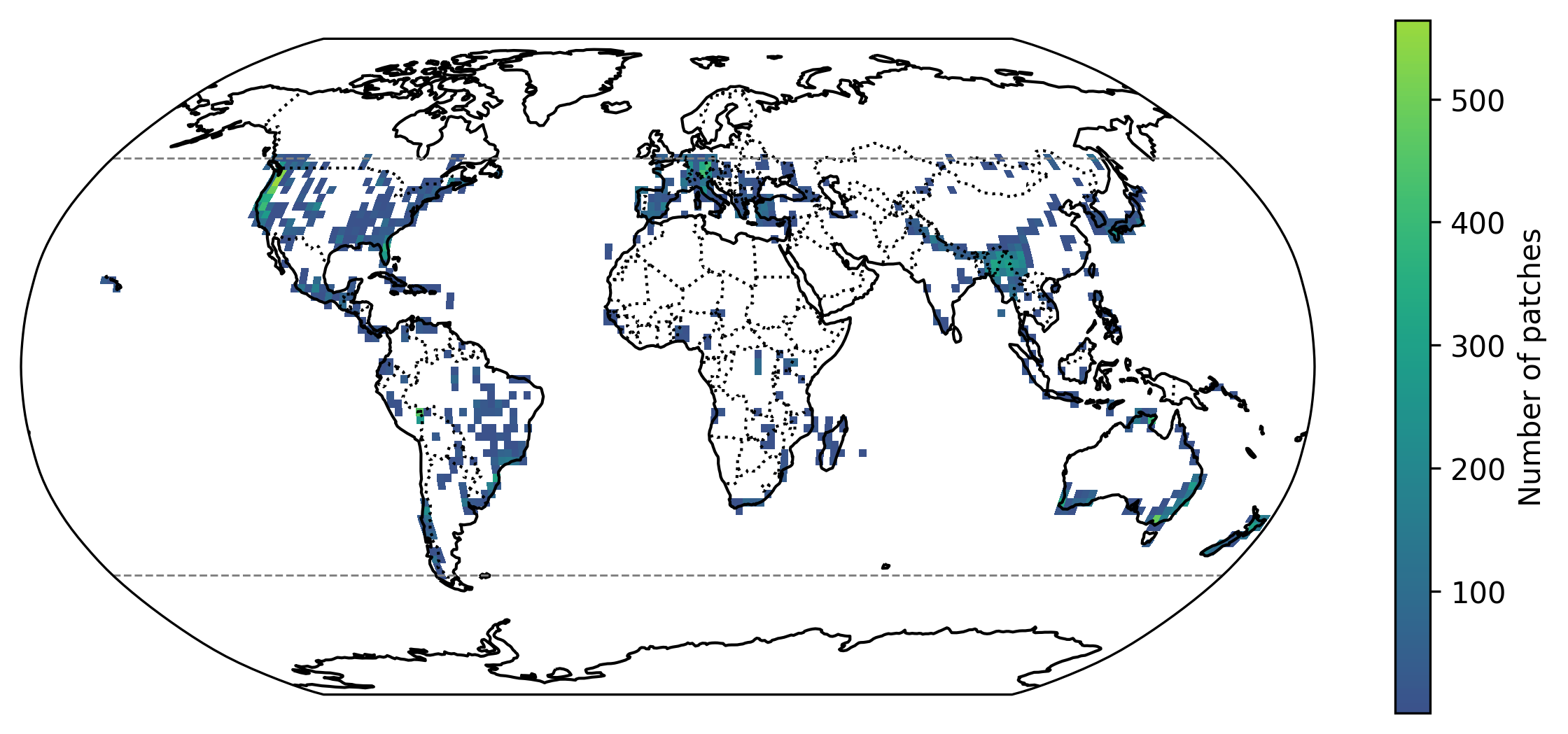}
\vspace{-0.6cm}
\caption{Geographic distribution of patches in the final dataset. Dotted lines at 51.6 degrees N and S indicate the spatial bounds of the GEDI mission.}
\label{fig:heatmap_patches}
\vspace{-.3cm}
\end{figure}

\begin{figure}[!t]
\centering
\includegraphics[width=\linewidth]{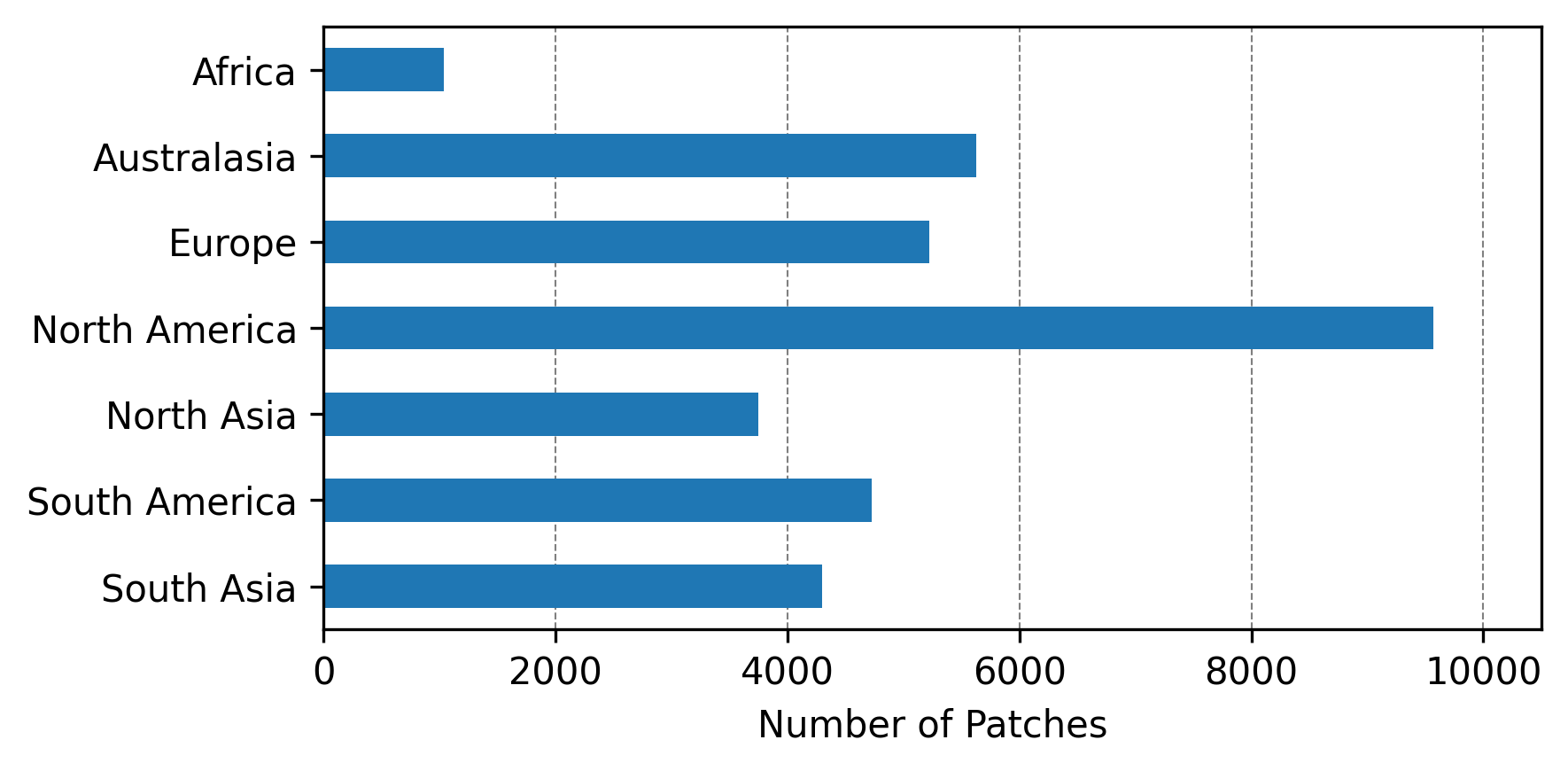}
\vspace{-.8cm}
\caption{Number of patches per continental world region.}
\label{fig:patch_statistics}
\vspace{-.3cm}
\end{figure}

\begin{figure}[!t]
\centering
\includegraphics[width=\linewidth]{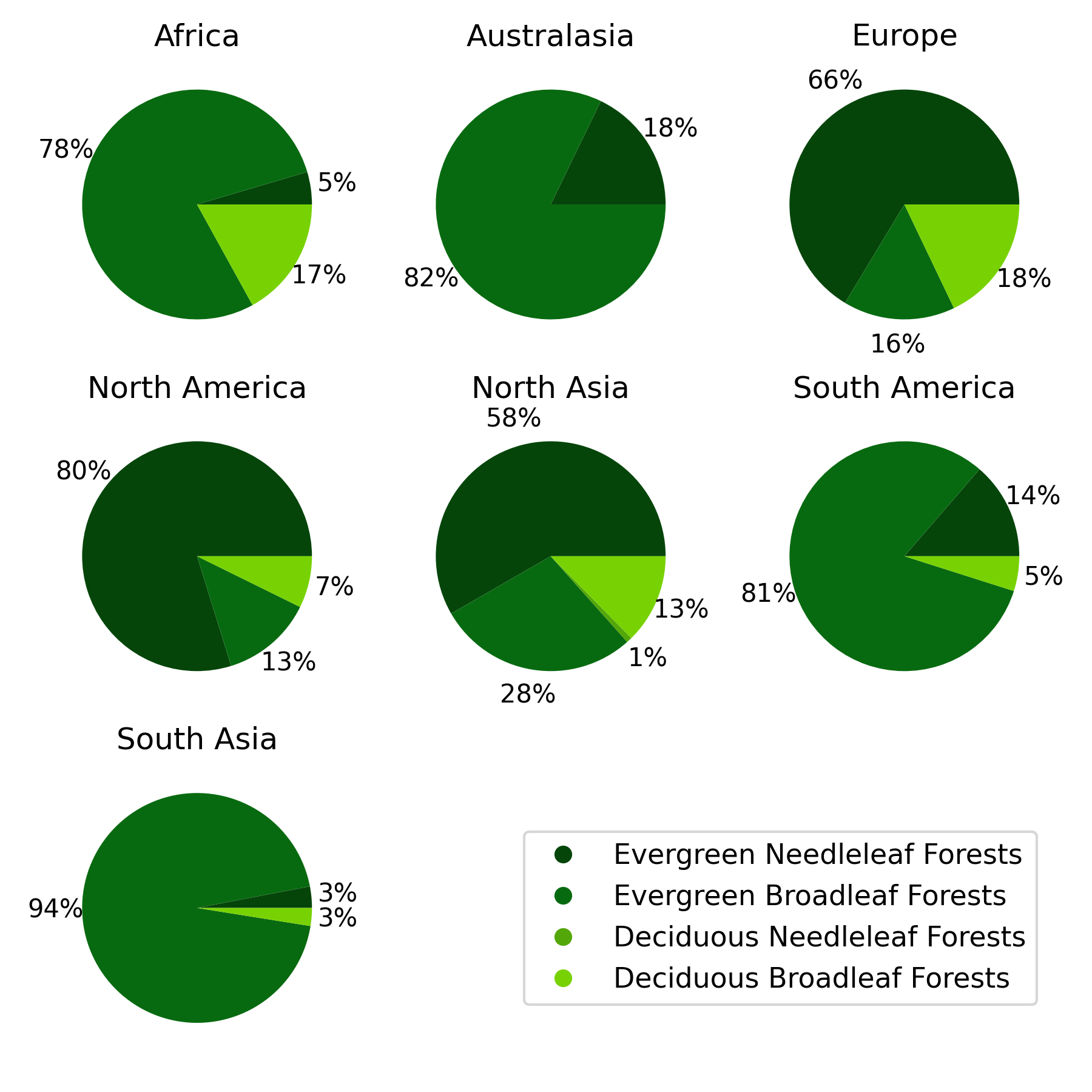}
\vspace{-1cm}
\caption{Distribution of plant functional types (PFTs) across the seven continental regions. The percentages are based on the fraction of GEDI shots corresponding to each PFT for the given region.}
\label{fig:pft_distribution_regions}
\vspace{-0.3cm}
\end{figure}

The geographic distribution of patches in the dataset is illustrated in \autoref{fig:heatmap_patches}, which is primarily determined by the distribution of forest areas and the availability of nearly cloud-free EnMAP tiles.
The fact that EnMAP is a German mission based on user's requests explains for example the over-representation of patches over Germany compared to France.
This results in a relatively even distribution of the number of patches per region with the exception of North America (9570 patches) and Africa (1041 patches), as visualized in \autoref{fig:patch_statistics}. The distribution of PFTs per region is shown in \autoref{fig:pft_distribution_regions}, highlighting the uniqueness of each region and revealing similarities between regions, such as Europe and North America.

\section{Experimental Setup for Benchmarking Geo-FMs}

\begin{table*}[!htbp]
    \centering
    \caption{Benchmark Results Across Regions for the Baseline U-Net and Pre-trained Foundation Models for Frozen Encoder and Fine-tuned Encoder. The Reported Mean and Standard Deviation Is Calculated Based on Five Random Dataset Splits for Each Region. The Reported Average Is the Mean of the Performance Per Region. The Best Score for Each Protocol and Region Is Highlighted in \textbf{Bold}, the Second Best Is \underline{Underlined}.}
    \vspace{-0.2cm}
    \label{tab:results_table}
    \begin{adjustbox}{width=\textwidth}
    \begin{tabular}{@{}llcccccccccc@{}}
        \toprule
        \multirow{2}{*}{\textbf{Region}} & \phantom{} & \multicolumn{2}{c}{\textbf{U-Net}} & \multicolumn{2}{c}{\textbf{Spectral-ResNet-50}} & \multicolumn{2}{c}{\textbf{Spectral-ViT-B}} & \multicolumn{2}{c}{\textbf{DOFA ViT-B}} & \multicolumn{2}{c}{\textbf{Panopticon}} \\
        \cmidrule(lr){3-4}\cmidrule(lr){5-6}\cmidrule(lr){7-8}\cmidrule(lr){9-10}\cmidrule(lr){11-12}
         & & RMSE & $R^2$ & RMSE & $R^2$ & RMSE & $R^2$ & RMSE & $R^2$ & RMSE & $R^2$ \\
        \midrule
        \multicolumn{2}{c}{} & \multicolumn{2}{c}{\multirow{-1}{*}{\textbf{(Supervised Baseline)}}} & \multicolumn{8}{c}{\textbf{Frozen Encoder}} \\
        \cmidrule(lr){3-4}
        \cmidrule(lr){5-12}
        \textbf{Africa} & & 65.65 ± 2.55 & 0.56 ± 0.03 & \second{61.87 ± 2.43} & \second{0.61 ± 0.03} & \best{60.83 ± 2.28} & \best{0.62 ± 0.03} & 66.02 ± 1.69 & 0.55 ± 0.04 & 63.04 ± 1.8 & 0.6 ± 0.03 \\
        \textbf{Australasia} & & 87.95 ± 1.39 & 0.47 ± 0.02 & \second{85.76 ± 1.71} & \second{0.49 ± 0.01} & \best{83.9 ± 1.66} & \best{0.51 ± 0.01} & 90.66 ± 1.43 & 0.43 ± 0.01 & 87.01 ± 1.7 & 0.48 ± 0.01 \\
        \textbf{Europe} & & \second{74.29 ± 1.15} & \second{0.38 ± 0.02} & 74.37 ± 1.13 & 0.38 ± 0.02 & \best{72.71 ± 1.17} & \best{0.41 ± 0.01} & 78.65 ± 1.29 & 0.31 ± 0.02 & 76.59 ± 1.33 & 0.34 ± 0.01 \\
        \textbf{North America} & & \second{90.62 ± 0.88} & \second{0.41 ± 0.01} & \best{90.45 ± 0.96} & \best{0.41 ± 0.01} & 91.19 ± 1.25 & 0.4 ± 0.02 & 96.23 ± 1.1 & 0.33 ± 0.01 & 91.07 ± 0.87 & 0.4 ± 0.01 \\
        \textbf{North Asia} & & 94.67 ± 0.85 & 0.3 ± 0.01 & \second{94.01 ± 0.68} & \second{0.31 ± 0.01} & \best{91.78 ± 1.42} & \best{0.34 ± 0.02} & 97.35 ± 0.7 & 0.26 ± 0.01 & 94.36 ± 0.93 & 0.3 ± 0.01 \\
        \textbf{South America} & & \second{82.79 ± 1.95} & \second{0.49 ± 0.02} & \best{81.79 ± 1.93} & \best{0.5 ± 0.02} & 83.26 ± 2.38 & 0.48 ± 0.03 & 86.12 ± 1.71 & 0.45 ± 0.02 & 83.15 ± 1.94 & 0.48 ± 0.02 \\
        \textbf{South Asia} & & \best{97.34 ± 0.84} & \best{0.27 ± 0.01} & \second{97.95 ± 0.52} & \second{0.26 ± 0.01} & 98.07 ± 1.53 & 0.26 ± 0.03 & 100.5 ± 0.48 & 0.22 ± 0.01 & 98.83 ± 0.86 & 0.25 ± 0.01 \\
        \cmidrule(lr){3-12}
        \textbf{Average} & & 84.76 & 0.41 & \second{83.74} & \second{0.42} & \best{83.11} & \best{0.43} & 87.93 & 0.36 & 84.86 & 0.41 \\
        \midrule
        \multicolumn{2}{c}{} & \multicolumn{2}{c}{\multirow{-1}{*}{\textbf{(Supervised Baseline)}}} & \multicolumn{8}{c}{\textbf{Fine-tuned Encoder}} \\
        \cmidrule(lr){3-4}
        \cmidrule(lr){5-12}
        \textbf{Africa} & & 65.65 ± 2.55 & 0.56 ± 0.03 & \best{59.04 ± 2.35} & \best{0.65 ± 0.03} & \second{62.34 ± 1.67} & \second{0.6 ± 0.03} & 67.47 ± 1.51 & 0.54 ± 0.04 & 65.82 ± 2.1 & 0.56 ± 0.03 \\
        \textbf{Australasia} & & 87.95 ± 1.39 & 0.47 ± 0.02 & \best{80.08 ± 1.6} & \best{0.56 ± 0.01} & \second{80.57 ± 2.01} & \second{0.55 ± 0.01} & 88.92 ± 2.27 & 0.45 ± 0.02 & 85.2 ± 1.82 & 0.5 ± 0.01 \\
        \textbf{Europe} & & 74.29 ± 1.15 & 0.38 ± 0.02 & \second{69.6 ± 1.17} & \second{0.46 ± 0.01} & \best{69.08 ± 1.25} & \best{0.46 ± 0.01} & 77.25 ± 1.28 & 0.33 ± 0.02 & 74.98 ± 1.43 & 0.37 ± 0.01 \\
        \textbf{North America} & & 90.62 ± 0.88 & 0.41 ± 0.01 & \second{82.45 ± 0.86} & \second{0.51 ± 0.01} & \best{81.94 ± 0.76} & \best{0.52 ± 0.01} & 92.53 ± 0.8 & 0.38 ± 0.01 & 88.75 ± 1.0 & 0.43 ± 0.01 \\
        \textbf{North Asia} & & 94.67 ± 0.85 & 0.3 ± 0.01 & \best{89.23 ± 0.91} & \best{0.38 ± 0.01} & \second{89.28 ± 1.06} & \second{0.38 ± 0.01} & 99.43 ± 0.62 & 0.23 ± 0.01 & 95.5 ± 1.17 & 0.29 ± 0.01 \\
        \textbf{South America} & & 82.79 ± 1.95 & 0.49 ± 0.02 & \best{77.6 ± 1.99} & \best{0.55 ± 0.02} & \second{78.77 ± 1.93} & \second{0.54 ± 0.02} & 84.34 ± 1.79 & 0.47 ± 0.02 & 82.51 ± 1.91 & 0.49 ± 0.02 \\
        \textbf{South Asia} & & 97.34 ± 0.84 & 0.27 ± 0.01 & \best{94.03 ± 0.97} & \best{0.32 ± 0.01} & \second{94.06 ± 0.66} & \second{0.32 ± 0.01} & 99.71 ± 0.79 & 0.23 ± 0.02 & 98.58 ± 0.59 & 0.25 ± 0.01 \\
        \cmidrule(lr){3-12}
        \textbf{Average} & & 84.76 & 0.41 & \best{78.86} & \best{0.49} & \second{79.43} & \second{0.48} & 87.09 & 0.38 & 84.48 & 0.41 \\
        \bottomrule
    \end{tabular}
    \end{adjustbox}
\end{table*}

We evaluate the performance of all Geo-FMs under two settings: (i) using frozen encoders with pre-trained weights to assess the quality of extracted features, and (ii) fully fine-tuning including the encoder. For all Vision Transformer (ViT)-based GeoFMs we implement a UPerNet \cite{upernet} decoder, following the setup of existing benchmarks like PANGAEA \cite{pangaea}. All Geo-FMs included in this benchmark are based on the ViT-B architecture with 12 layers, where we select the 4th, 6th, 8th and 12th intermediate feature layers to serve as input to the UPerNet. The output is then upsampled to the original patch size using bilinear interpolation. 
For the Spectral-ResNet-50, we append two final convolutional layers as a simple decoder, following the original publication's setting \cite{spectralearth}.
While different pre-training strategies have been compared in SpectralEarth \cite{spectralearth}, we limit our experiments to two variants: \mbox{Spectral-ResNet-50}, pre-trained with \mbox{MoCo-V2} \cite{mocov2}, and Spectral-ViT-B, pre-trained using the Masked Autoencoder (MAE) approach \cite{mae}.
Whereas DOFA has also been trained using the MAE methodology, Panopticon is built on the \mbox{DINOv2} framework \cite{dinov2}.
To provide a baseline for comparison, we additionally train a fully supervised U-Net \cite{unet}, which has demonstrated impressive results in many tasks, including AGB estimation \cite{unet_pascarella}.
We limit hyperparameter tuning to the learning rate of the AdamW optimizer and train all models for 100~epochs using mean squared error loss with early stopping.
We train all models using five seeds, resulting in distinct dataset splits, with a 70:20:10 split ratio for training, validation, and testing in the regional dataset.
Since DOFA and Panopticon were pre-trained on 224$\times$224 pixel patches, we resample the input data and labels using bilinear interpolation to match the pre-training dimensions of these models.

\section{Benchmarking Results}
The benchmarking results are presented in \autoref{tab:results_table}.
In the frozen encoder setting, the average $R^2$ values across all regions range from 0.36 for DOFA ViT-B to 0.43 for Spectral-ViT-B. The \mbox{U-Net's} performance falls in between, varying by dataset size per region. Specifically, for North America, the \mbox{U-Net's} performance is comparable to that of the top-performing \mbox{Geo-FM}, whereas for Africa, which has the smallest dataset, the gap is significantly larger.
The performance achievable in each region varies significantly, likely due to the differing complexities of estimating AGBD in each region and the accuracy of AGBD estimates used as labels.
In the fine-tuned encoder setting, Spectral-ResNet50 and Spectral-ViT-B exhibit a substantial performance improvement, outperforming the \mbox{U-Net} across all regions.
Interestingly, both DOFA and Panopticon exhibit superior performance in the fine-tuned encoder setting than in the frozen encoder setting across all regions, except Africa and North Asia. This deviation may be attributed to increased data requirements associated with fine-tuning these models. DOFA's average performance improvement can also be explained by the limited amount of HSI data in the pre-training dataset, making the benefits of fine-tuning the encoder more pronounced.

\begin{figure*}[!t]
\centering
\includegraphics[width=\linewidth]{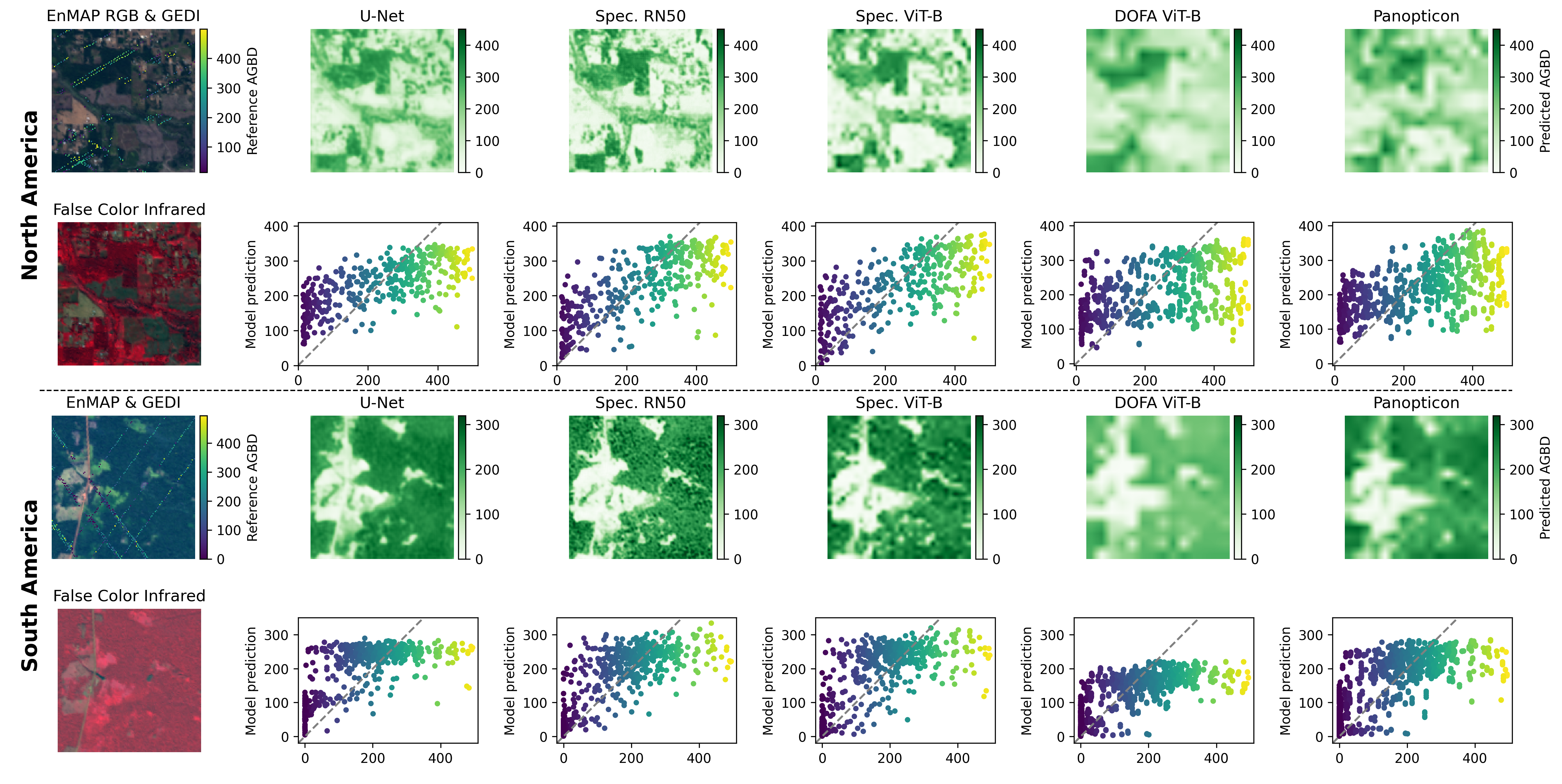}
\vspace{-0.7cm}
\caption{Model predictions for one patch from North America (top two rows) and another patch from South America (bottom two rows). Each EnMAP patch is visualized in RGB with overlaid GEDI reference AGBD and as infrared false color composite. The scatter plots show the relationship between reference AGBD values and predictions for each model. Predictions are generated using Geo-FMs with a fine-tuned encoder.}
\label{models_predictions}
\end{figure*}

For a qualitative evaluation of the results, we present the predictions of each model for a sample patch from North America and South America in \autoref{models_predictions}. The EnMAP patch is visualized as an infrared false color composite and as RGB, with the reference AGBD values from the GEDI shots overlaid to highlight the sparse nature of the labels. The predictions are generated by Geo-FMs with a fully fine-tuned encoder for the corresponding region. 
Visually, the U-Net, Spectral-\mbox{ResNet-50}, and Spectral-ViT-B produce the highest quality results. In contrast, predictions by DOFA and Panopticon appear blurry, which can be explained by their token patch sizes of 16 and 14, respectively.
To address this, previous studies have implemented pixel shuffle layers instead of bilinear interpolation of the UPerNet output for a learned pixel-wise reconstruction \cite{muszynski}. To improve the pixel-level reconstruction quality in the evaluation of Prthvi 2.0 on pixel-wise AGB regression, the architecture was modified to Prithvi-UNet, which integrates the pre-trained Prithvi encoder and decoder as the bottleneck in a U-Net architecture, tuning all model parameters \cite{prithvi2}.
It is important to note that the varying token patch sizes in the ViT-based Geo-FMs hinder direct comparisons on their approaches to handling HSI, as this has a significant impact on the performance.
The scatter plots in \autoref{models_predictions} visualize the relationship between the labels (x-axis) and AGB predictions (y-axis) by each model for all the GEDI shots in the patch. Across all models, an overestimation of low-reference AGB values and an underestimation of high-reference AGB values are observed. This bias has been consistently demonstrated in AGB maps produced by remote sensing models and can be attributed to signal saturation, leading to an underestimation of high AGB values, among other sources of error \cite{rejou‑mechain}.

\section{Conclusion}
This letter presents a novel globally-distributed hyperspectral benchmark dataset for forest AGB estimation.
Our benchmarking results demonstrate that existing Geo-FMs achieve comparable performance to the baseline U-Net when the encoder is frozen, and significant improvements are observed when fully fine-tuning the encoder, particularly for Spectral-ResNet-50 and Spectral-ViT, highlighting their potential for hyperspectral biomass mapping. We also demonstrate that the token patch size is a key design factor in ViT backbones for pixel-wise regression tasks.
By releasing this dataset, we aim to facilitate the development and evaluation of Geo-FMs for HSI, and enable systematically studying the geographic bias, generalization capacity, and the design of pretraining strategies to improve transferability in global remote sensing applications.

\vfill

\end{document}